\documentclass[10pt]{article}
\usepackage[margin=1in]{geometry}
\pdfoutput=1

\usepackage[colorlinks,citecolor=red]{hyperref}

\usepackage{graphicx}
\DeclareGraphicsExtensions{.pdf}
\usepackage[justification=Centering,singlelinecheck=false,caption=false,font=footnotesize]{subfig}

\usepackage{natbib}
\bibpunct{(}{)}{;}{a}{}{,}

\usepackage[cmex10]{amsmath}
\usepackage{amsfonts}
\usepackage{bm}
\numberwithin{equation}{section}

\usepackage{amsmath}
\usepackage{amsfonts}
\usepackage{amssymb}
\usepackage{color}

\usepackage{algorithm,algpseudocode}
\newtheorem{definition}{Definition}

\usepackage{graphicx}
\usepackage{url}
\usepackage{algorithm}
\usepackage{subfig}
\usepackage{array}

\usepackage{multirow}
\usepackage{array}
\renewcommand{\arraystretch}{1.5}
\usepackage{times}
\usepackage{xcolor}
\usepackage{soul}
\usepackage[utf8]{inputenc}
\usepackage{bbding}
\usepackage{makecell}

\usepackage{graphicx,float}
\usepackage{cases}
\usepackage{bm}
\usepackage{booktabs}
\usepackage{enumerate}

\usepackage{times}
\usepackage{xcolor}
\usepackage{soul}
\usepackage[utf8]{inputenc}
\usepackage{bbding}
\usepackage{makecell}
\usepackage[small]{caption}
\usepackage{amsmath,amssymb,mathrsfs}
\usepackage{graphicx,float}
\usepackage{cases}
\usepackage{multirow,array}
\usepackage{bm}
\usepackage{color}
\usepackage{array}
\usepackage{booktabs}
\usepackage{enumerate}
\usepackage{enumitem}
\usepackage{url}

\usepackage{stmaryrd}
\usepackage{dsfont}

\usepackage{algpseudocode}

\def\ie{\emph{i.e.}}
\def\eg{\emph{e.g.}}

\begin{document}

\title{\textbf{Multi-class Label Noise Learning via\\ Loss Decomposition and Centroid Estimation}}

\author{Yongliang Ding \hspace{1cm} Tao Zhou 
\hspace{1cm} Chuang Zhang\\
 Yijing Luo
 \hspace{1.5cm}Juan Tang
 \hspace{1cm}Chen Gong \\
PCA Lab, the Key Laboratory of Intelligent Perceptron and Systems for High-Dimensional Information \\of Ministry of Education, School of Computer Science and Engineering, \\Nanjing University of Science and Technology, \\China Jiangsu Key Lab of Image and Video Understanding for Social Security\\School of Computer Science and Cyber \\Engineering, Guangzhou University, China\\
\{ylding; c.zhang; yjluo; chen.gong\}@njust.edu.cn, taozhou.ai@gmail.com, tanjn16@gzhu.edu.cn}

\date{20 December 2021}

\maketitle

\begin{abstract}
In real-world scenarios, many large-scale datasets often contain inaccurate labels, \ie, noisy labels, which may confuse model training and lead to performance degradation. To overcome this issue, Label Noise Learning (LNL) has recently attracted much attention, and various methods have been proposed to design an unbiased risk estimator to the noise-free dataset to combat such label noise. Among them, a trend of works based on Loss Decomposition and Centroid Estimation (LDCE) has shown very promising performance. However, existing LNL methods based on LDCE are only designed for binary classification, and they are not directly extendable to multi-class situations. In this paper, we propose a novel multi-class robust learning method for LDCE, which is termed ``MC-LDCE''. Specifically, we decompose the commonly adopted loss (\eg, mean squared loss) function into a label-dependent part and a label-independent part, in which only the former is influenced by label noise. Further, by defining a new form of data centroid, we transform the recovery problem of a label-dependent part to a centroid estimation problem. Finally, by critically examining the mathematical expectation of clean data centroid given the observed noisy set, the centroid can be estimated which helps to build an unbiased risk estimator for multi-class learning. The proposed MC-LDCE method is general and applicable to different types (\ie, linear and nonlinear) of classification models. The experimental results on five public datasets demonstrate the superiority of the proposed MC-LDCE against other representative LNL methods in tackling multi-class label noise problem.
\textbf{Keywords:} 
Multi-class Classification; Label Noise; Loss Decomposition; Centroid Estimation. 
\end{abstract}

\section{Introduction}
\noindent With the availability of large quantities of well-annotated datasets, recent learning methods such as deep learning have been proven successful in various practical tasks. However, it is very expensive to collect large-scale well-annotated data in some fields, such as medical image analysis, speech translation, natural language processing, and so on. Currently, several strategies with low cost have been developed to collect labeled data for many tasks, such as automatic web crawlers and crowdsourcing. However, these strategies inevitably introduce many incorrect labels, due to the limitation of technologies and human expertise, leading to dramatic performance degradation of learning models ~\cite{ChangFraudDetection,WangTackling}. Therefore, developing effective Label Noise Learning (LNL) algorithms is highly demanded in various real-world applications. 

Up to now, different LNL methods have been proposed to deal with the label noise problem \cite{han2020survey} over the last few years, and they can be roughly divided into three categories. The first category is sample selection ~\cite{kumar2010self,jiang2015self,han2018co} which relies on the memorization effect of neural networks to select probable correctly-labeled examples. To be specific, neural networks tend to overfit the small-loss examples which are considered as clean data in the early learning stage, then gradually overfit the large-loss examples which are likely to be contaminated by label noise. Therefore, many methods based on choosing small-loss examples are proposed to improve LNL performance. The second category is label correction ~\cite{brodley1999identifying,reed2014training,tanaka2018joint} which attempts to identify and correct the potentially incorrect labels through joint optimization of label purification and network weights. 
The third category is loss correction ~\cite{natarajan2013learning,ghosh2015making,van2015learning} which modifies the loss functions to be further minimized and makes them robust to noisy labels.

Among the above three categories of methods, loss correction has shown very promising performance due to a solid mathematical foundation, and a popular way to convert the conventional losses to the robust ones is based on loss decomposition and centroid estimation (LDCE), such as Labeled Instance Centroid Smoothing (LICS) \cite{gao2016risk}, $\mu$SGD \cite{patrini2016loss}, and Centroid Estimation with Guaranteed Efficiency (CEGE) \cite{gong2020centroid}. These methods aim to estimate the real data centroid under a clean set using the observed noisy data so that a noise-robust loss function can be obtained. However, the above works can only deal with binary classification tasks and can hardly be applied to multi-class cases. The reasons are two-fold. First, they only focus on decomposing the binary classification loss such as hinge loss and perceptron loss. As a result, it is difficult to apply these models to multi-class classification tasks. Second, they need to use the facts that $y^2=1$ for the label $y=+1$ or $-1$ and the positive label $+1$ and negative label $-1$ only differ in the sign. Unfortunately, these facts do not hold under multi-class cases anymore. Therefore, developing LNL methods based on LDCE for the multi-class classification task is highly demanded.
 
To this end, we propose a new \textbf{M}ulti-\textbf{C}lass LNL method via loss decomposition and \textbf{C}entroid \textbf{E}stimation (termed MC-LDCE) to deal with LNL problems. Specifically, we propose to decompose the multi-class classification loss (\eg, mean squared loss) into a label-independent part and a label-dependent part, so that the multi-class label noise only affects the second part. Then by defining a new form of data centroid, we observe that the label-dependent part is strongly related to such centroid which critically governs the model robustness. Further, by investigating the mathematical expectation of centroid under the noisy dataset as well as introducing an elementary row transformation matrix, such data centroid can be estimated which leads to an unbiased risk estimator to the noise-free case for multi-class learning. Besides, our MC-LDCE is quite general and independent of the classification models, which does not need auxiliary clean data unlike some existing methods such as \cite{ren2018learning} and \cite{veit2017learning}. The experimental results on typical benchmarks and real-world noisy datasets show that MC-LDCE outperforms the representative LNL methods under different types of multi-class label noise.

\section{Related Work}
LNL is an important branch of weakly-supervised learning ~\cite{ChenGong2021,Chuang2019,Gong2019, Chuang2020} which has attracted intensive research over the past decades.
We review three major types of existing LNL methods, including sample selection, label correction, and loss correction. 
\par 
\textbf{Sample selection}. The methods of this type try to select the correctly-labeled examples according to different criteria. For example, Jiang~\emph{et al.} \cite{jiang2018mentornet} proposed MentorNet to teach another student network to select the examples with probably correct labels during training. However, such a selection method cannot overcome the inferiority of accumulated error caused by sample-selection bias. To overcome such drawbacks, Han~\emph{et al.} \cite{han2018co} proposed Co-teaching to train two networks simultaneously and update itself with the data selected by its peer network. As for Co-teaching+ \cite{yu2019does}, it improves Co-teaching by only selecting the small-loss instances with different predictions from two networks. To further explore the information inherited from data, Wei~\emph{et al.} \cite{wei2020combating} proposed to use a joint loss to select small-loss examples, so that more data with the consensus of two networks can be selected. 
 
\par
\textbf{Label correction}. Label correction is a quite intuitive solution that identifies the possible incorrectly-labeled data and then corrects their labels for reliable training \cite{brodley1999identifying}. However, such clean data identification and correction can be imprecise. Therefore, Samel~\emph{et al.} \cite{samel2018active} presented a new active deep denoising approach that first builds a deep neural network noise model and then applies an active learning algorithm to identify the optimal denoising function. Besides, Sheng~\emph{et al.} \cite{sheng2017majority} effectively correct the labels when the data are collected from the crowdsourcing platform. The works of \cite{reed2014training,hu2019weakly} proposed a self-training scheme that alternatively conducts label correction and trains a deep neural network on noisily-labeled data. However, these methods use original noisy labels for learning properly, so the performance is still likely to be degraded by the noisy data. Therefore, \cite{tanaka2018joint} takes a similar way to \cite{reed2014training}, but it completely replaces all labels with pseudo-labels for model training. Similarly, Song~\emph{et al.} \cite{song2019selfie} proposed a method of which the key idea is to selectively refurbish and exploit unclean samples that can be corrected with high precision, which results in an augmentation of examples available for training. 

\par
\textbf{Loss correction}. This type of LNL methods aim to correct the traditional loss functions to make them robust to label noise. Natarajan~\emph{et al.} \cite{natarajan2013learning} provided the general form of a noise corrected loss $\hat{l}$, making the minimization of $\hat{l}$ is equivalent to the minimization of the noise-free loss $l$ on clean data given the noise rate. To a further step, if the noise rate is not available, Ghosh~\emph{et al.} \cite{ghosh2015making} provided a condition to be satisfied when $l$ is robust to label noise. Van~\emph{et al.} \cite{van2015learning} provided several examples of such robust non-convex losses, while Masnadi~\emph{et al.} \cite{masnadi2008design} showed that the linear or unhinged loss itself is noise-corrected loss. Besides, Jindal~\emph{et al.} \cite{jindal2016learning} augmented the standard network by adding a softmax layer for estimating the confusion matrix. Jin~\emph{et al.} \cite{jin2021pattern} proposed a maximum likelihood estimation based objective function for robust classification.

\section{Preliminaries}
    In this section, we first introduce some notations which will be used in this paper. In traditional supervised learning, we define the $\mathcal{X} \in \mathbb{R}^d $ and $\mathcal{Y} = \{ 0,1 \}^c$ as the input feature space and the output label space, where $d$ denotes the dimension of features and $c$ denotes the number of classes. For a matrix $\mathbf{M}$, we define $\mathbf{M^\dagger}$ as its pseudo-inverse matrix and $\mathbf{M}^\top$ as its transpose. 
    The superscript ``$\widetilde{\quad}$'' stands for the variable being noisy or estimated and the ``$\hat{\quad}$'' indicates that the variable is an empirical quantity.

    Let $\mathcal{D}$ be the underlying noise-free joint distribution of a pair of random variables $(X,Y)\in \mathcal{X} \times \mathcal{Y} $. The clean training set $S=\{{(\mathbf{x}_i,\mathbf{y}_i)}\}_{i=1}^{n}$ of $(X,Y)$ containing $n$ i.i.d. data points can be drawn identically and independently from $\mathcal{D}$, where $\{\mathbf{x}_i\}_{i=1}^{n}$ are the feature representations of $n$ examples and $\{\mathbf{y}_i\}_{i=1}^{n}$ are one-hot label vectors which are correct. However, for the supervised learning with noisy labels, we only have access to the noisy distribution $\widetilde{\mathcal{D}}$ of random variables $(X,\widetilde{Y})\in \mathcal{X} \times \mathcal{{Y}}$, and the noisy training set $\widetilde{S}=\{{(\mathbf{x}_i,\widetilde{\mathbf{y}}_i)}\}_{i=1}^{n}$ containing $n$ i.i.d. data points which is drawn from a noisy distribution $\widetilde{\mathcal{D}}$. In our learning problem under noisy labels, the target is to find a suitable multi-class decision function $h \in \mathcal{H}: \mathcal{X} \rightarrow \mathcal{Y}$ with parameter matrix $\mathbf{W} \in \mathbb{R}^{d \times c}$ training on $\widetilde{S}$, where $\mathcal{H}$ represents the hypothesis space, such that $h$ can precisely predict the label $\mathbf{y}$ of any test example $\mathbf{x} \in \mathcal{X}$. 

	To further understand the corruption process under label noise, we introduce a noise transition matrix $\mathbf{T} \in \left[0,1 \right]^{c \times c}$, of which the main purpose is to model the transition from latent clean label $Y$ to the observed noisy label $\widetilde{Y}$. In this case, the element $\mathbf{T}_{ij}=p(\widetilde{Y}=\mathbf{e}_j|Y=\mathbf{e}_i)$ stands for the probability of clean label $\mathbf{e}_i$ wrongly annotated as $\mathbf{e}_j$ where $\mathbf{e}_i$ denotes the one-hot vector with the $i$-th element being 1 and the others being 0.

	\section{Proposed Method}
    In this section, we will detail our proposed MC-LDCE algorithm.
	\subsection{Multi-class Loss Decomposition}\label{section_loss_decomposition}
	\hspace*{\fill} \\
	\noindent
	We define $l: \mathbb{R} \times \mathcal{Y} \rightarrow \mathbb{R}$ as the loss function that penalizes the difference between the model output $h(\mathbf{x})$ and the ground-truth label $\mathbf{y}$ under traditional supervised learning. Thus, the empirical risk on a clean set $S$ can be formulated as
	\begin{equation}\label{empiricalOnS}
	\hat{\mathcal{R}}(h,S)=\frac{1}{n}\sum\limits_{i=1}^{n}\ell(h(\mathbf{x}_i),\mathbf{y}_i),
	\end{equation}
	where $h$ is the shorthand for $h(\mathbf{x})$ in this paper if no confusion is incurred. Similar to Eq.~\eqref{empiricalOnS}, we can define the empirical risk on corrupted data set $\widetilde{S}$ as Eq.~\eqref{eq:EmpiricalRiskOnSTilde}.
	\begin{equation}\label{eq:EmpiricalRiskOnSTilde}
	\widetilde{\mathcal{R}}(h,\widetilde{S})=\frac{1}{n}\sum\limits_{i=1}^{n}\ell(h(\mathbf{x}_i),\widetilde{\mathbf{y}}_i).
	\end{equation} 
	
	Because of the unavailability of the ground-truth $\{\mathbf{y}_i\}_{i=1}^{n}$, the $\widetilde{\mathcal{R}}(h,\widetilde{S})$ can be deviated from the real $\hat{\mathcal{R}}(h,S)$.
	It is expected to find an unbiased estimator $\mathcal{\widetilde{\hat{R}}}(h,\widetilde{S})$ for $\mathcal{\hat{R}}(h,S)$ on the noisy set $\widetilde{S}$, so that the negative impact caused by the noisy label can be eliminated. 
	
	As mentioned earlier, our method is based on LDCE to deal with multi-class LNL, while previous works ~\cite{patrini2016loss,gong2020centroid} based on LDCE are designed to solve the binary LNL. By decomposing the mean squared loss and expressing the decision function as $h(\mathbf{x};\mathbf{W})=\left\langle\mathbf{W},\mathbf{x}\right\rangle$. Eq.~\eqref{empiricalOnS} can be reformulated as:
	\begin{flalign}\label{MSELoss_decomposition}
	&\hat{\mathcal{R}}(h,S)\\
	&=\frac{1}{n}\sum_{i=1}^{n}\|\mathbf{y}_i-\mathbf{W}^{\top}\mathbf{x}_i \|_{2}^2  \nonumber\\
	&=\frac{1}{n}\sum_{i=1}^{n} \left( \mathbf{y}_i^{\top}\mathbf{y}_i-2\mathbf{y}_i^{\top}\mathbf{W}^{\top}\mathbf{x}_i+ \mathbf{x}_i^{\top}\mathbf{W}\mathbf{W}^{\top}\mathbf{x}_i   \right) \nonumber \\
	&=\frac{1}{n} \sum_{i=1}^{n} \left(\mathbf{y}_i^{\top}\mathbf{y}_i+\mathbf{x}_i^{\top}\mathbf{WW}^{\top}\mathbf{x}_i \right)- \frac{2}{n}\sum_{i=1}^{n}\mathbf{y}_i^{\top}\mathbf{W}^{\top}\mathbf{x}_i \nonumber.
	\end{flalign}
	
	It should be noted that when the label vector $\mathbf{y}_i$ follows the form of one-hot encoding, $\mathbf{y}_{i}^{\top}\mathbf{y}_i=1$ always holds. Besides, from the knowledge of linear algebra, the following equation holds, which is
	\begin{equation}
	\mathbf{y}_i^{\top}\mathbf{W}^{\top}\mathbf{x}_i=\mathrm{trace}(\mathbf{y}_i^{\top}\mathbf{W}^{\top}\mathbf{x}_i)=\mathrm{trace}(\mathbf{W}^{\top}\mathbf{x}_i\mathbf{y}_i^{\top}).
	\label{eq5}
	\end{equation}
	
	Therefore, according to Eq.~\eqref{eq5}, Eq.~\eqref{MSELoss_decomposition} can be derived as 
	\begin{flalign}
	\label{eq_trace}
	&\hat{\mathcal{R}}(h,S) \\
	&=\frac{1}{n} \sum_{i=1}^{n} \left(1+\mathbf{x}_i^{\top}\mathbf{WW}^{\top}\mathbf{x}_i \right)- \frac{2}{n}\sum_{i=1}^{n}\mathrm{trace}(\mathbf{W}^{\top}\mathbf{x}_i \mathbf{y}_i^{\top}) \nonumber \\
	&=(1+\frac{1}{n} \sum_{i=1}^{n} \mathbf{x}_i^{\top}\mathbf{WW}^{\top}\mathbf{x}_i) -\frac{2}{n}\mathrm{trace}( \mathbf{W}^{\top}\sum_{i=1}^{n} \mathbf{x}_i\mathbf{y}_i^{\top} ) \nonumber \\
	&=(1+\frac{1}{n} \sum_{i=1}^{n} \mathbf{x}_i^{\top}\mathbf{WW}^{\top}\mathbf{x}_i) -2\mathrm{trace}( \mathbf{W}^{\top} \hat{\mu}(S) ) \nonumber,
	\end{flalign}
	where the empirical centroid $\hat{\mu}(S)$ of the clean data set $S$ is defined as
	\begin{equation}\label{empirical_centroid}
	    \hat{\mu}(S)=\frac{1}{n}\sum_{i=1}^{n}\mathbf{x}_i\mathbf{y}_i^{\top}.
	\end{equation}
	
	Note that the dataset centroid $\hat{\mu}(S)$ defined here is different from of that defined in previous works~\cite{patrini2016loss,gong2020centroid} which target binary classification, and this is essential for our method to handling multi-class classification. Corresponding to Eq~\eqref{empirical_centroid}, the mathematical expectation of the centroid on the entire distribution $\mathcal{D}$ is defined as 
	\begin{equation}\label{mathematical_expectation}
	  \mu(\mathcal{D})=\mathbb{E}_{(X,Y)\sim \mathcal{D}}[XY^{\top}],
	\end{equation}
	where $\mathbb{E}[\cdot]$ calculates the mathematical expectation.
	From Eq.~\eqref{eq_trace}, we see that only the second term is related to the label value $\mathbf{y}_i$. Therefore, if we want to find an unbiased $\widetilde{\hat{\mathcal{R}}}(h,\widetilde{S})$ to $\hat{\mathcal{R}}(h,S)$ to deal with noisy labels, we need to accurately estimate the dataset centroid $\hat{\mu}({S})$ based on $\widetilde{S}$, and the obtained estimator for $\hat{\mu}({S})$ is dubbed $\widetilde{\hat{\mu}}({S})$. 
	Finally, the proposed MC-LDCE model can be built by combining the empirical risk Eq.~\eqref{eq_trace} with some techniques for preventing overfittings, such as the linear model with $\ell_2$ regularizer, and the neural network with a dropout operation.\par
    From Eq.~\eqref{eq_trace}, we see the key to recover the empirical risk on clean set is to accurately estimate the centroid (\ie, Eq.~\eqref{empirical_centroid}), and this will be detailed in the following section. 
	\subsection{Centroid Estimation}\label{section_centroid}
	\hspace*{\fill} \\
	\noindent
We estimate $\hat{\mu}(S)$ of the clean set $S$ via the centroid $\hat{\mu}(\widetilde{S})$ of the noisy set $\widetilde{S}$. To this end, we investigate the mathematical expectation of the centroid on the noisy set, which can be formulated by
    \begin{flalign}\label{eq:E_YX_SP}
	&\mathbb{E}_{\widetilde{Y}}[X\widetilde{Y}^{\top}|(X,Y)]=\sum_{i=1}^{c}   \pi_i\mathbb{E}_{\widetilde{Y}}[X\widetilde{Y}^{\top}|(X,Y=\mathbf{e}_{i})], 
	\end{flalign}
	where $\mathbf{e}_i$ represents a one-hot vector of which only the $i$-th element is 1, so $Y=\mathbf{e}_i$ denotes that the example belongs to the $i$-th class. Besides, $\pi_i=P(Y=\mathbf{e}_i)$ stands for the prior probability of the $i$-th class. To compute the value of Eq.~\eqref{eq:E_YX_SP}, we need the following definition:
\begin{definition}\label{definition_matrix}
	{\rm Suppose there are two one-hot label vectors, where $\mathbf{y}_i$ has the $i$-th element being 1, and $\mathbf{y}_j$ has the $j$-th element being 1 where $i \neq j$. Then the two vectors can be converted using an \emph{imputation matrix} $\mathbf{K}_{i \rightarrow j}$, which is
	\begin{equation}
	\mathbf{y}_j=\mathbf{K}_{i \rightarrow j}\mathbf{y}_i,
	\end{equation}
	where $\mathbf{K}_{i \rightarrow j}$ can be obtained by swapping the $i$-th row and the $j$-th row of an identity matrix $\mathbf{I}$.}
\end{definition}

	Therefore, for the $j$-th class, we introduce the \emph{imputation matrix} $\mathbf{K}_{i \rightarrow j}$ defined in Definition~\ref{definition_matrix}, and then have
\begin{flalign}\label{estimation_for_ith_class}
	&\mathbb{E}_{\widetilde{Y}}[X\widetilde{Y}^{\top}|(X,Y=\mathbf{e}_i)] =  \sum_{j=1}^{c}\mathbf{T}_{ij}{XY}^{\top}\mathbf{K}_{i \rightarrow j}^{\top},
\end{flalign}
where $\mathbf{T}$ is the noise transition matrix defined in Section 3. The estimation for this matrix can be completed by employing some off-the-shelf tools such as the method in ~\cite{liu2015classification,xia2019anchor}. In this work, we use the state-of-art method VolMinNet \cite{li2021provably} to estimate $\mathbf{T}$. 

	\renewcommand\arraystretch{0.3}	
	\begin{algorithm}[t]
		\caption{The overall algorithm of MC-LDCE.}
		\label{algorithm}
		\begin{algorithmic}[1]
			\State{\bf Input:} Noisy training dataset $\widetilde{S}=\{(\mathbf{x}_i,\widetilde{\mathbf{y}}_i)\}_{i=1}^n$;
			\State
		    Estimate the transition matrix $\textbf{T}$ via VolMinNet \cite{li2021provably};
		   \State Compute all class priors $\pi_1, \cdots, \pi_c$ via Eq.~\eqref{eq_priors};
		   \State Compute $\mathbf{M}$ via Eq.~\eqref{calculation_M};
		   \State Compute the estimated centroid of $S$ via Eq.~\eqref{estiation_with_matrix};
		   \State Compute the unbiased risk estimator $\widetilde{\hat{\mathcal{R}}}(h,\widetilde{S})$ via Eq.~\eqref{final_loss};
		   \State Use any off-the-shelf solver to optimize the model (\eg, linear model or CNN) by employing the $\widetilde{\hat{\mathcal{R}}}(h,\widetilde{S})$ as the loss function;  
		   \State \textbf{Output:} Optimal parameters $\mathbf{W}$.
		\end{algorithmic}
	\end{algorithm}

Based on Eq.~\eqref{estimation_for_ith_class}, Eq.~\eqref{eq:E_YX_SP} can be further derived as 
\begin{flalign}
\label{calculation_M}
	\mathbb{E}_{\widetilde{Y}}[X\widetilde{Y}^{\top}|(X,Y)] 
	&\quad =\sum_{i=1}^{c} \pi_i\sum_{j=1}^{c}\mathbf{T}_{ij}X{Y}^{\top}\mathbf{K}_{i \rightarrow j}^{\top}  \\
	 &\quad = X{Y}^{\top} \underbrace {\left[ \sum_{i=1}^{c}\pi_i \sum_{j=1}^{c}\mathbf{T}_{ij}\mathbf{K}_{i \rightarrow j}^{\top}   \right]}_{\mathbf{M}} \nonumber.
\end{flalign}

Here we denote $\mathbf{M}=\sum_{i=1}^{c}\pi_i \sum_{j=1}^{c}\mathbf{T}_{ij}\mathbf{K}_{i \rightarrow j}^{\top}$. Thus, the unbiased estimator $\widetilde{\hat{\mu}}(S)$ can be formulated as
\begin{flalign}\label{estiation_with_matrix}
	&\widetilde{\hat{\mu}}(S)= \hat{\mu}(\widetilde{S} )\mathbf{M}^{\dagger},
\end{flalign} 
where the $\mathbf{M}^{\dagger}$ stands for the pseudo inverse matrix of $\mathbf{M}$.

Finally, the unbiased risk estimator $\widetilde{\hat{\mathcal{R}}}(h,\widetilde{S})$ to $\hat{\mathcal{R}}(h,S)$ under noisy set $\widetilde{S}$ can be obtained by substituting Eq.~\eqref{estiation_with_matrix} to Eq.~\eqref{eq_trace}, which can be represented as 
\begin{equation}\label{final_loss}
    \begin{aligned}
    \widetilde{\hat{\mathcal{R}}}(h,\widetilde{S}) =  1+\frac{1}{n} \sum_{i=1}^{n}\mathbf{x}_i^{\top}\mathbf{WW}^{\top}\mathbf{x}_i -2\mathrm{trace}(\mathbf{W}\hat{\mu}(\widetilde{S})\mathbf{M}^{\dagger}).
    \end{aligned}
\end{equation}

\subsection{Class Prior Estimation}\label{section_class_prior_estimation}
\hspace*{\fill} \\
\noindent 
Note that in Eq.~\eqref{calculation_M}, it needs to obtain the class prior $\pi_1, \cdots, \pi_c$, thus we present how to estimate them based on the noise transition matrix $\mathbf{T}$ in this subsection, and we will describe the corruption process from clean labels to noisy labels. The element $\mathbf{T}_{ij}=P(\widetilde{Y}=\mathbf{e}_j|Y=\mathbf{e}_i)$ in the matrix represents the label flip rate from the $i$-th class to the $j$-th class as defined before. Obviously, $\sum_{j=1}^{c}\mathbf{T}_{ij}=1$. The class priors are defined as $\pi_1=P(Y=\mathbf{e}_1), \pi_2=P(Y=\mathbf{e}_2), \cdots, \pi_c=P(Y=\mathbf{e}_c)$ which can be easily obtained by solving the following equation \begin{equation}\label{eq_priors} 
\small
\left\{ \begin{aligned}
    &P(\widetilde{Y}=\mathbf{e}_1)=\mathbf{T}_{11}\pi_{1}+\mathbf{T}_{21}\pi_{2}+\cdots+\mathbf{T}_{c1}\pi_{c}\\
    &~~~~~~~~~~~~~~~~~~~~~~~~~~~~~\vdots\\
    &P(\widetilde{Y}=\mathbf{e}_i)=\mathbf{T}_{1i}\pi_{1}+\mathbf{T}_{2i}\pi_{2}+\cdots+\mathbf{T}_{ci}\pi_{c}\ \ \   ,\\
    &~~~~~~~~~~~~~~~~~~~~~~~~~~~~~\vdots\\
    &P(\widetilde{Y}=\mathbf{e}_c)=\mathbf{T}_{1c}\pi_{1}+\mathbf{T}_{2c}\pi_{2}+\cdots+\mathbf{T}_{cc}\pi_{c} 
    \end{aligned}
    \right.
\end{equation}
where the left-hand side of every equation can be directly estimated based on the noisy $\widetilde{S}$.

\subsection{Summary of the Proposed Method}
\hspace*{\fill} \\
\noindent
From Subsection~\ref{section_loss_decomposition} to Subsection~\ref{section_class_prior_estimation}, we see that the proposed MC-LDCE approach decomposes the multi-class classification loss (\eg,{ mean squared loss}) into a label-independent part and a label-dependent part, and then directly estimates the label-dependent part via centroid estimation, which makes it can solve the multi-class LNL problem. It is worth noting that MC-LDCE is a simple yet effective LNL algorithm, which is flexible and compatible with different types of classification models $h(\mathbf{x})$ (\eg, deep nonlinear models and linear models). 
The detailed steps of our method are summarized in \textbf{Algorithm} \ref{algorithm}.

\begin{table}[t]\centering
\small
	\caption{The characteristic of \textit{CIFAR-10}, \textit{MNIST}, \textit{FASHION-MNIST}, \textit{SVHN} and \textit{Animal-10N}.}\vspace{-0.25cm}
	\setlength\tabcolsep{8pt}
	 \renewcommand{\arraystretch}{1.0}
	\begin{tabular}{cccc} 
	\toprule[1pt]
		\textbf{Dataset}   & \textbf{\# train} & \textbf{\# test}  & \textbf{size }        \\ 
		\hline
		\textit{MNIST}             & 60,000      & 10,000               & 28$\times$28 \\ 
		\textit{FASHION-MNIST}     & 60,000      & 10,000              & 28$\times$28 \\ 
		\textit{CIFAR-10}          & 50,000      & 10,000              & 32$\times$32$\times$3 \\ 
		\textit{SVHN}              & 73,257      & 26,032             & 32$\times$32$\times$3 \\ 
		\textit{Animal-10N}        & 50,000      & 5,000        & 64$\times$64$\times$3 \\ 
	\toprule[1pt]
	\label{statistic_detail}
	\end{tabular} \vspace{-0.45cm}
\end{table}

\section{Experiments}

In this section, we first provide the experimental settings, including the characteristics of datasets, selected backbone, and some implementation details. Then, we present the experimental results on both simulated and real-world noisy datasets with deep classification models. Further, we validate the robustness of our MC-LDCE when a linear classification model is deployed. 

\subsection{Experiments with Deep Classification  Models} 
\hspace*{\fill} \\
In this part, we equip our MC-LDCE with deep classification models and compare it with several representative deep robust methods.
\subsubsection{Basic Setup}
\paragraph{Simulated Noisy Datasets.} 
We verify the effectiveness of our approach on four manually corrupted datasets (\ie, \textit{CIFAR-10}, \textit{MNIST}, \textit{FASHION-MNIST}, and \textit{SVHN}). The statistics of the used datasets are summarized in Table~\ref{statistic_detail}. Specifically, \textit{FASHION-MNIST} and \textit{MNIST} consist of $60,000$ images for training and $10,000$ images for testing, with the number of classes and the scale of each image being $10$ and $28\times28$, respectively. While \textit{CIFAR-10} and \textit{SVHN} are also considered $10$-class datasets, with the the scale of each image being $32\times32\times3$. \textit{CIFAR-10} contains $50,000$ training images and $10,000$ test images. \textit{SVHN} contains $73,257$ training images and $26,032$ test images. 
Note that all the original datasets are clean. Following the common setting in \cite{han2018co,yu2019does,wei2020combating}, we corrupted the training sets manually by using Sym-flipping and Pair-flipping noise transition matrices \cite{han2020survey}, with the noise rate being $\{20\%, 60\%\}$ and $\{20\%, 40\%\}$, respectively. To be specific, the Sym-flipping structure models the scenario where the class of clean label can uniformly flip into other classes, and the Pair-flipping structure models the situation where the class of a clean label can flip into its adjunct class instead of a far-away class.
\begin{table}[t]
    \center
    \small
    \setlength\tabcolsep{11pt}
    \caption{Network architectures of the adopted six-layer CNN and two-layer MLP.}\vspace{-0.25cm}
    \renewcommand{\arraystretch}{1.0}
    \begin{tabular}{cc}
        \toprule[1.0pt]
         six-layers CNN                   &    two-layers MLP      \\ 
         \hline
         3$\times$3 conv, 128 LReLU     &                  \\
         3$\times$3 conv, 128 LReLU     &                   \\
         3$\times$3 conv, 128 LReLU     &                   \\ 
         \cline{1-1}
        2$\times$2 max-pool, stride 2    &dense 784 $\rightarrow$ 256                   \\
        dropout, p = 0.25              &                   \\ 
        \cline{1-1}
         3$\times$3 conv, 512 LReLU    & 256 LReLU                  \\
         3$\times$3 conv, 256 LReLU    &                   \\
         3$\times$3 conv, 128 LReLU    &                   \\ 
         \cline{1-1}
         avg-pool                           &                   \\ 
         \hline{1-1}
         dense 128 $\rightarrow$ \#classes  & dense 256 $\rightarrow$ \#classes                  \\ 
     \toprule[1.0pt]
    \end{tabular}\label{tab:architecture}
\end{table}

\renewcommand\arraystretch{0.85}
	\begin{table*}[]
	\centering
    \renewcommand{\arraystretch}{0.52}
		\small
	\caption{Average test accuracy and the corresponding standard deviation on \emph{CIFAR-10}, \emph{MNIST}, \emph{FASHION-MNIST}, and \emph{SVHN} over the last ten epochs. The best results are marked in \textbf{bold}.} \vspace{-0.2cm}
	
			\begin{tabular}{c|c|c|c|c|c|c}
				\toprule[1.0pt]
				Dataset & Noise Type and Rate 					& GCE \cite{zhang2018generalized}	                 & Co-teaching+\cite{yu2019does} 		  & JoCoR \cite{wei2020combating} 		     &SIGUA  \cite{han2020sigua}      	  & MC-LDCE      \\ 
				\toprule[1.0pt]
					\multirow{4}*{\emph{CIFAR-10}} 
				& Symmetry-20$\%$								& 80.73$\pm$0.04        & 78.75$\pm$0.04  & 84.85$\pm$0.03 & 81.49$\pm$0.01 & \textbf{85.18$\pm$0.18} \\
				& Symmetry-60$\%$ 								& 57.40$\pm$0.08   		& 48.78$\pm$0.37  & 69.07$\pm$0.07 & 67.38$\pm$0.02 & \textbf{70.34$\pm$0.08} \\
				& Pairflip-20$\%$ 								& 79.03$\pm$0.03   		& 74.99$\pm$0.08  & 83.63$\pm$0.06 & 80.59$\pm$0.01 & \textbf{85.46$\pm$0.18} \\
				& Pairflip-40$\%$ 								& 60.01$\pm$0.05   		& 51.73$\pm$0.07  & 62.95$\pm$0.09 & 71.37$\pm$0.04 & \textbf{78.22$\pm$0.28} \\ \hline
				\multirow{4}*{\emph{MNIST}}& Symmetry-20$\%$ 	& 95.88$\pm$0.01        & 96.79$\pm$0.03  & 96.32$\pm$0.03 & 93.64$\pm$0.01 & \textbf{97.40$\pm$0.05} \\
				& Symmetry-60$\%$ 								& 93.95$\pm$0.01        & 93.83$\pm$0.07  & 94.10$\pm$0.06 & 84.39$\pm$0.05 & \textbf{94.84$\pm$0.02} \\
				& Pairflip-20$\%$ 								& 95.93$\pm$0.01        & 97.08$\pm$0.04  & 95.27$\pm$0.02 & 86.87$\pm$0.31 & \textbf{97.28$\pm$0.02} \\
				& Pairflip-40$\%$								& 95.16$\pm$0.01        & 91.57$\pm$0.09  & \textbf{95.52$\pm$0.07} & 73.72$\pm$0.25 & 93.06$\pm$0.19 \\ \hline
				\multirow{4}*{\emph{FASHION-MNIST}}
				& Symmetry-20$\%$ 								& 86.22$\pm$0.01        & 87.48$\pm$0.05  & 87.42$\pm$0.04 & 82.63$\pm$0.88 & \textbf{87.70$\pm$0.04} \\
				& Symmetry-60$\%$ 								& 84.27$\pm$0.01        & 76.64$\pm$0.03  & 83.92$\pm$0.08 & 75.12$\pm$0.22 & \textbf{84.70$\pm$0.04} \\
				& Pairflip-20$\%$ 								& 86.38$\pm$0.02        & 86.89$\pm$0.09  & 87.62$\pm$0.03 & 77.94$\pm$0.51 & \textbf{87.83$\pm$0.18} \\
				& Pairflip-40$\%$								& 85.18$\pm$0.02        & 69.10$\pm$0.06  & 82.90$\pm$0.06 & 70.55$\pm$6.17 & \textbf{85.38$\pm$0.36} \\ \hline
				\multirow{4}*{\emph{SVHN}}
				& Symmetry-20$\%$								& 81.29$\pm$0.01       	& 93.02$\pm$0.05  & 78.40$\pm$0.02 & 92.19$\pm$0.01 & \textbf{94.44$\pm$0.04} \\
				& Symmetry-60$\%$ 								& 56.24$\pm$0.01        & 72.13$\pm$0.27  & 36.49$\pm$0.05 & \textbf{82.85$\pm$0.01} & 77.76$\pm$0.15 \\
				& Pairflip-20$\%$ 								&   92.96$\pm$0.01   	& 92.55$\pm$0.03  & 77.15$\pm$0.02 & 90.81$\pm$0.01 & \textbf{93.35$\pm$0.04} \\
				& Pairflip-40$\%$ 							& 83.95$\pm$0.01& 72.49$\pm$0.09  & 54.96$\pm$0.03 & \textbf{88.97$\pm$0.01} & 77.74$\pm$0.18 \\ 
				\toprule[1pt]
			\end{tabular}
			\label{result_on_simulated_data}
		\end{table*}

\begin{figure*}[!h]
	\begin{minipage}{1\linewidth}
		\centering
		{\includegraphics[width=0.28\linewidth]{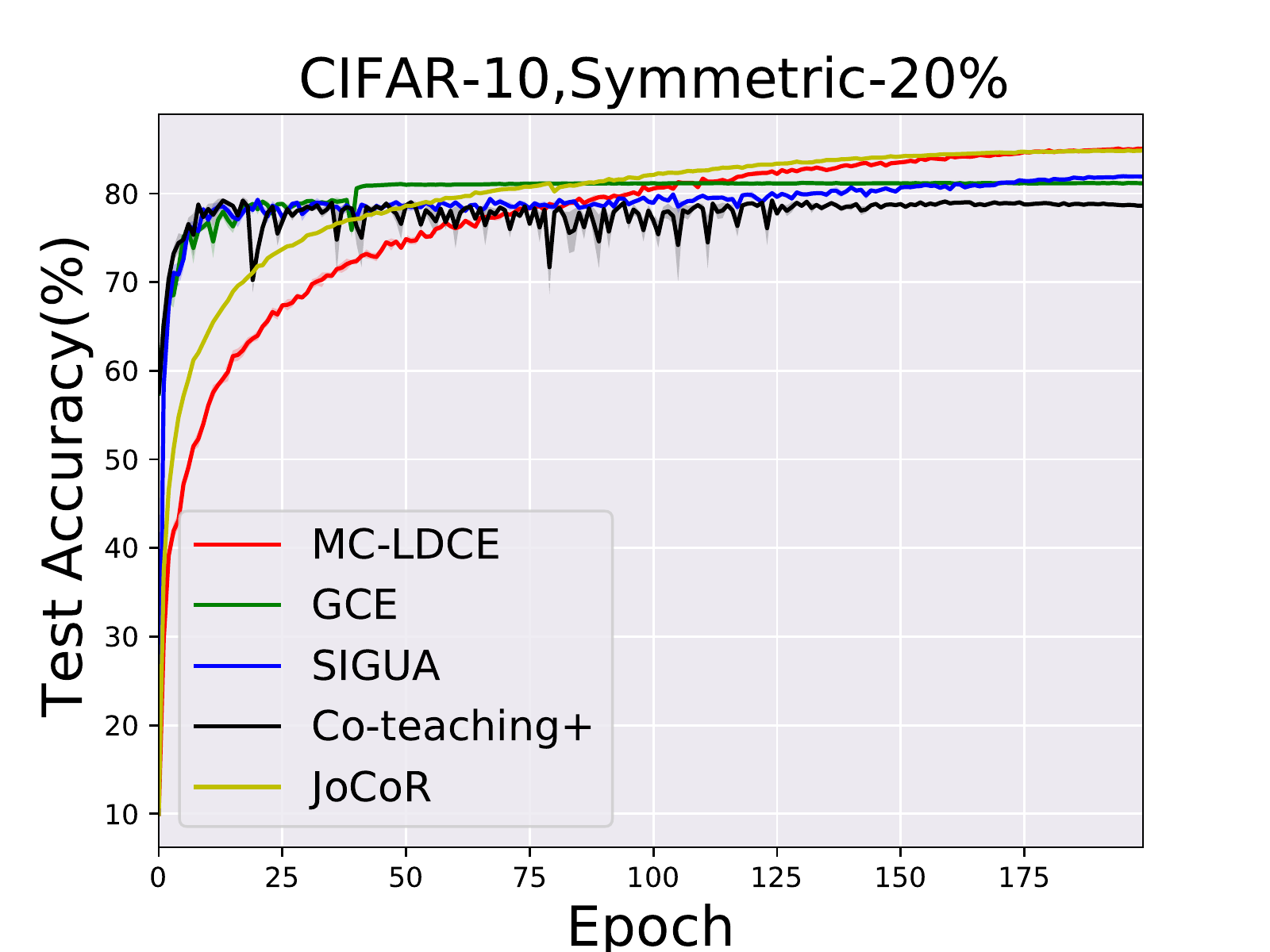}}		
		{\includegraphics[width=0.28\linewidth]{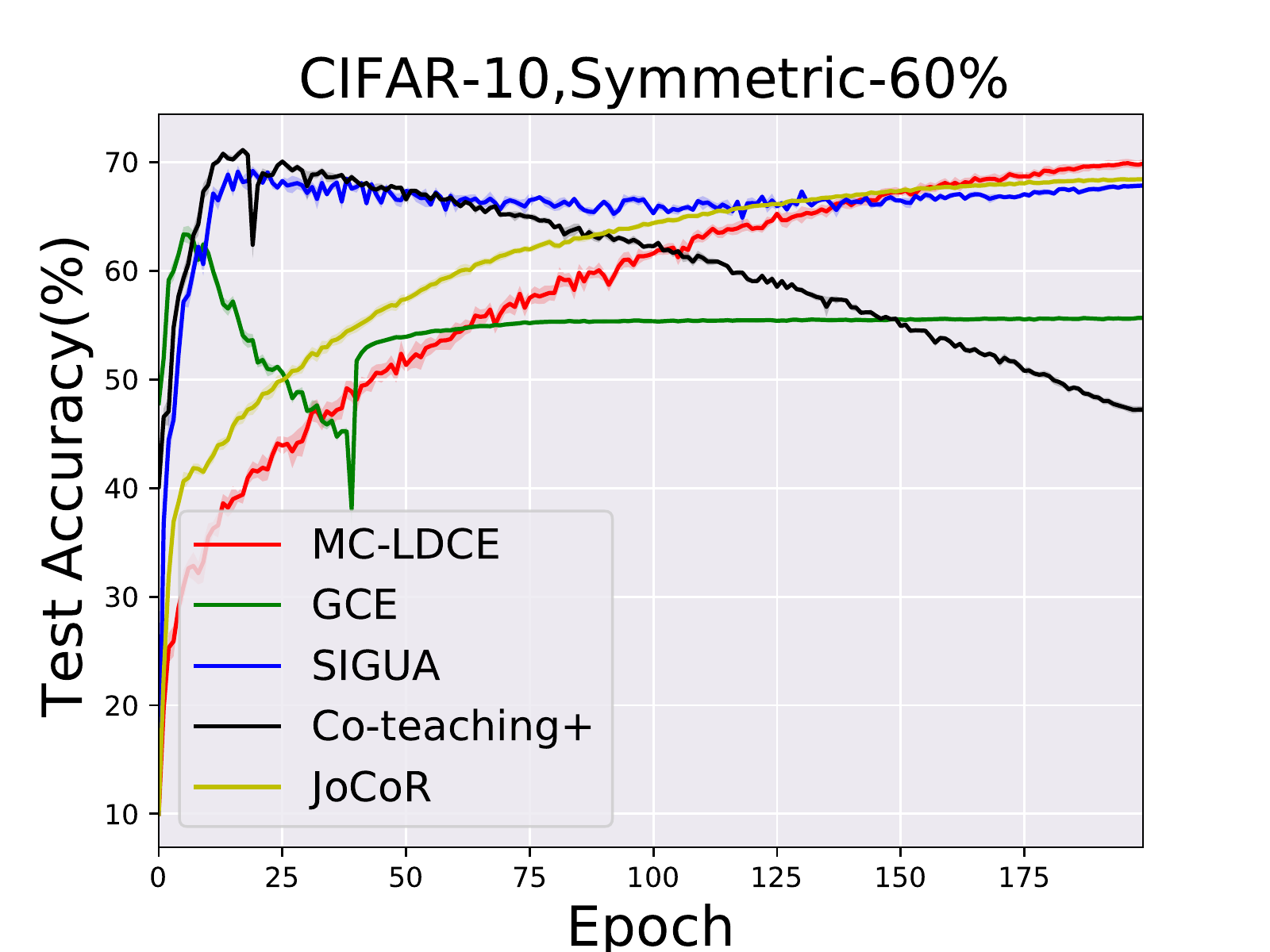}}
		{\includegraphics[width=0.28\linewidth]{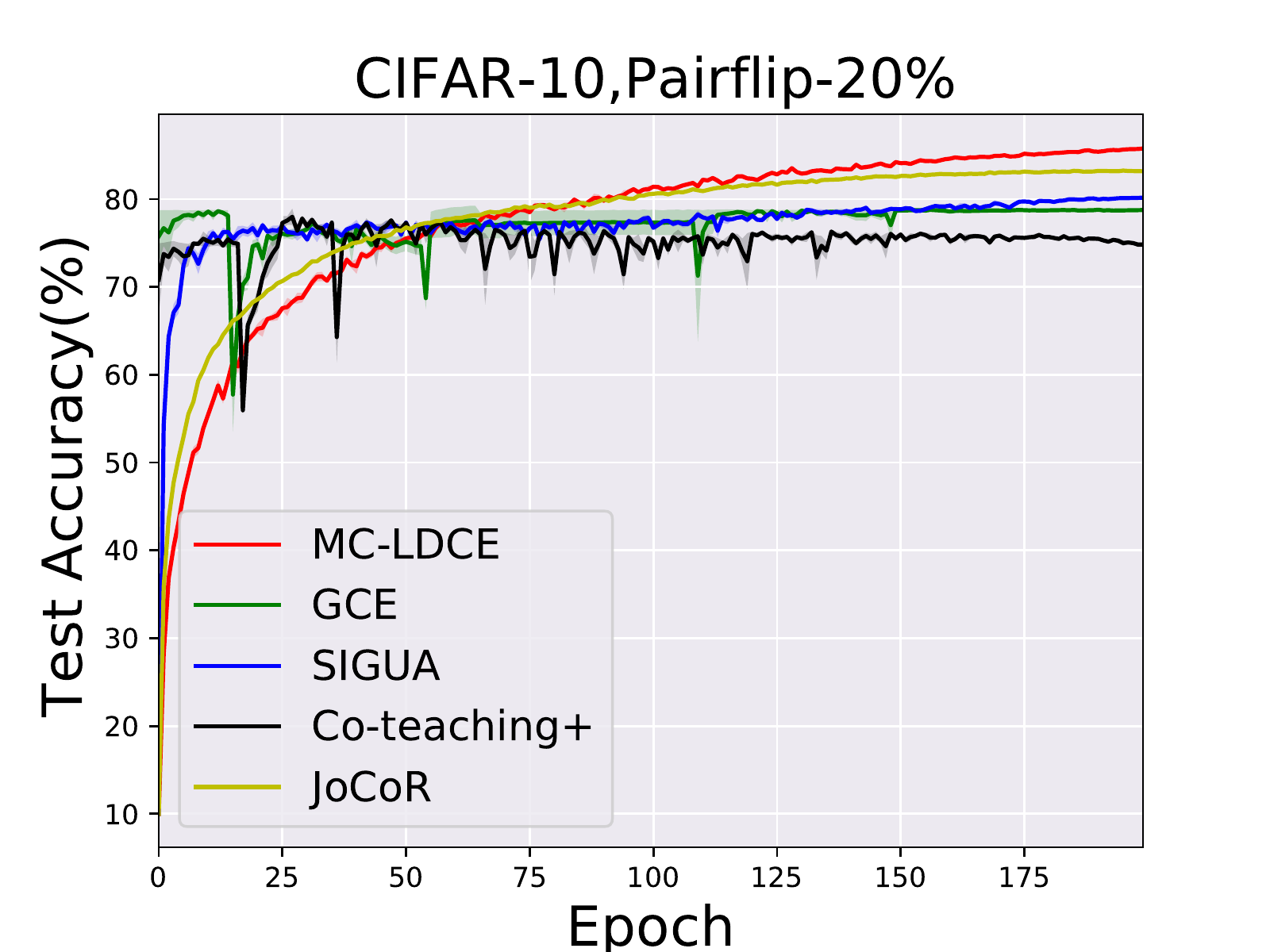}}
	\end{minipage} \vspace{-0.3cm}
	\caption{Test accuracy curves on \emph{CIFAR-10} with different noise rates for all the compared methods. Colored curves show the mean accuracy of five trials, and shaded bars denote the standard deviations of the accuracies over five trials. }\vspace{-0.3cm}
	\label{fig2}
\end{figure*}
\begin{figure*}[!h]
	\begin{minipage}{1\linewidth}
		\centering
		{\includegraphics[width=0.28\linewidth]{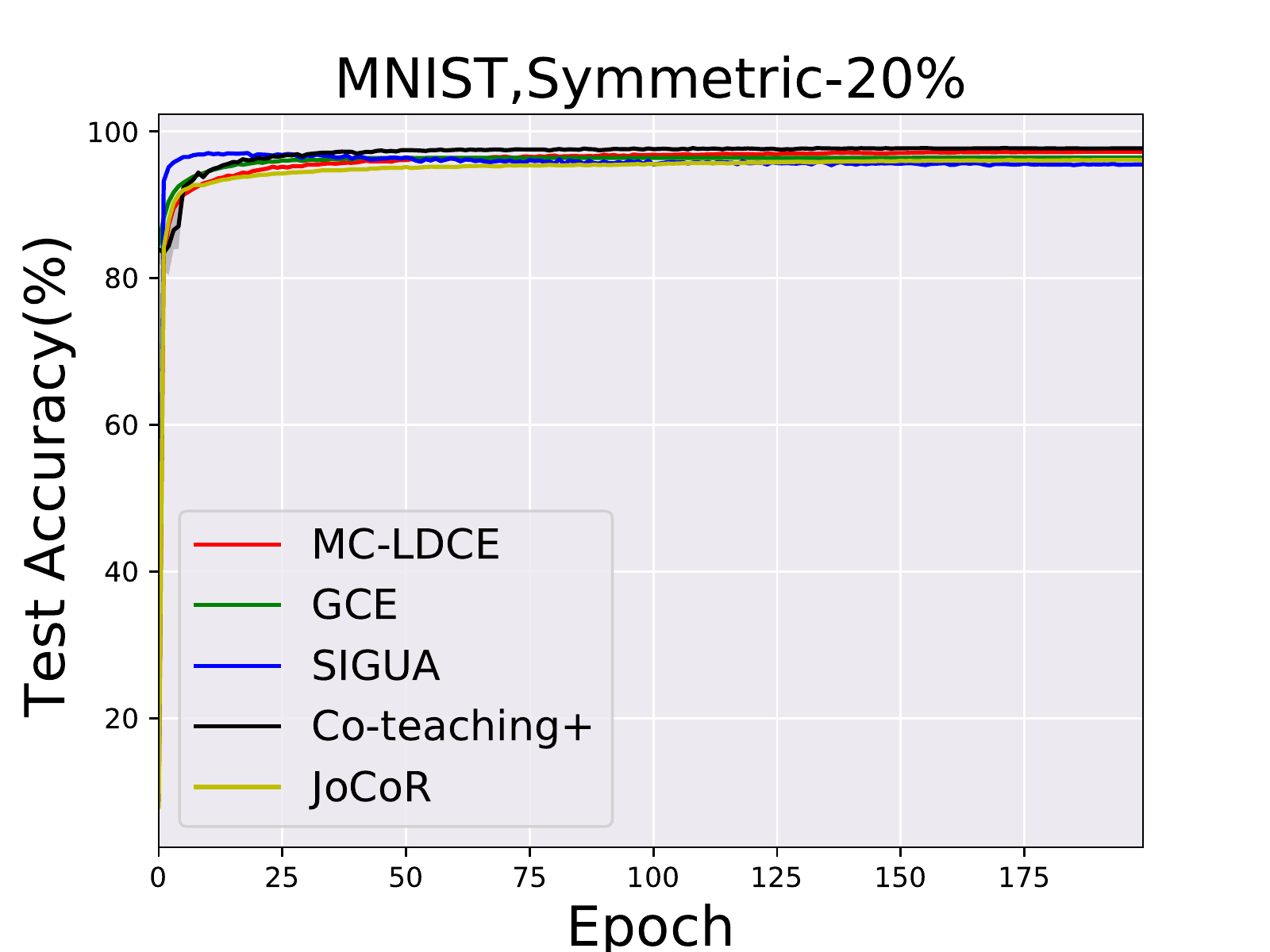}}		
		{\includegraphics[width=0.28\linewidth]{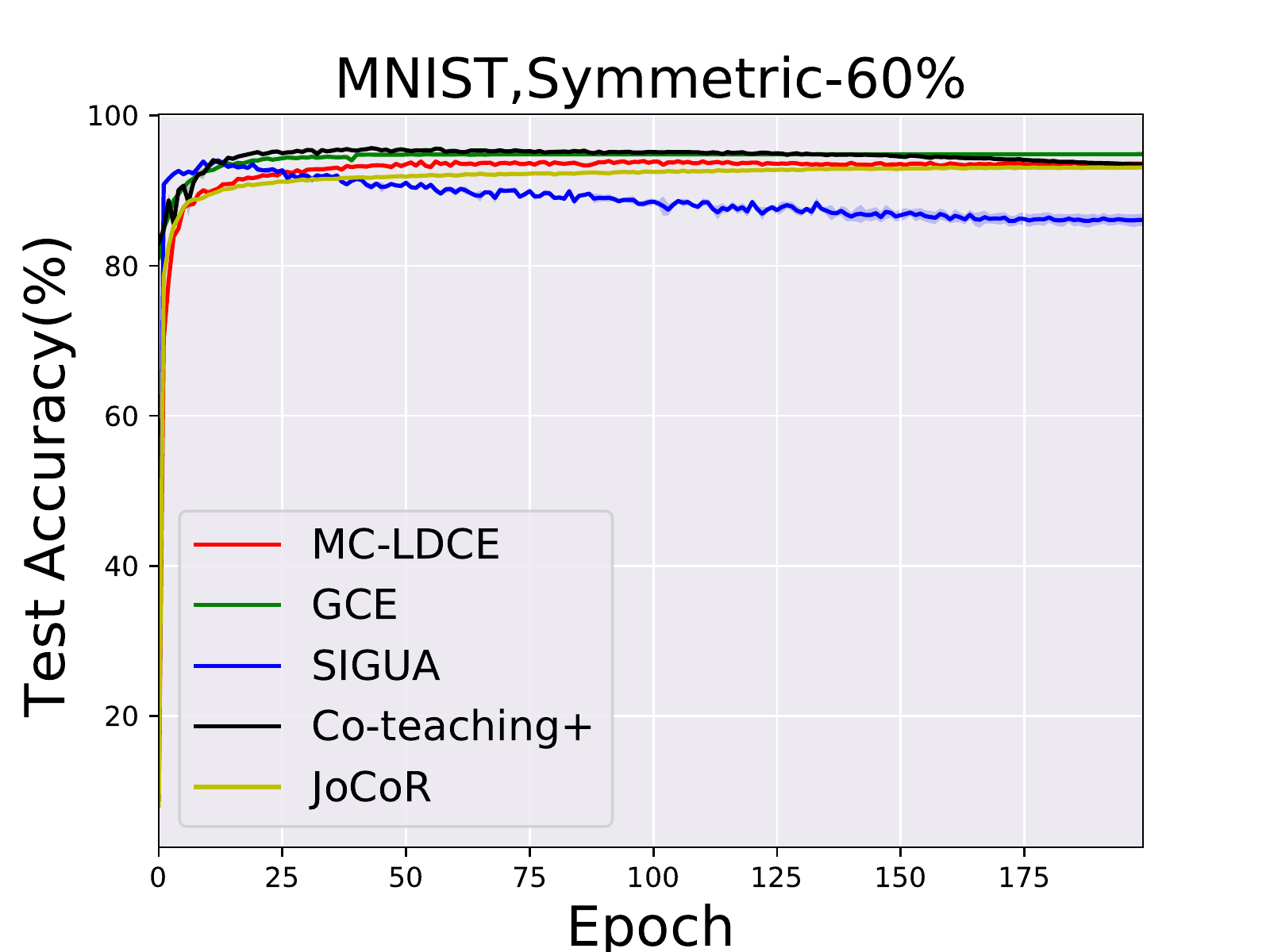}}
		{\includegraphics[width=0.28\linewidth]{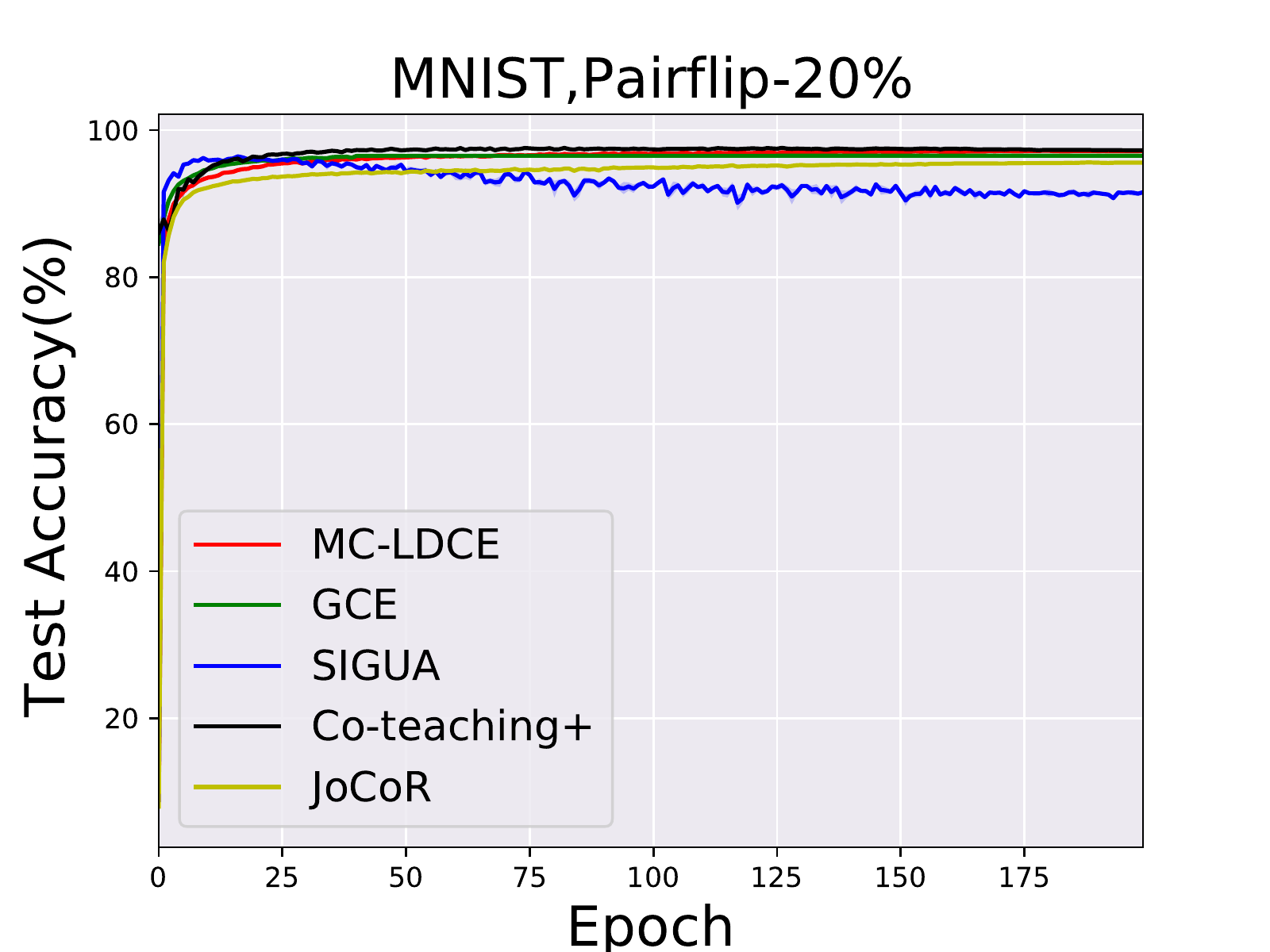}}
	\end{minipage} \vspace{-0.3cm}
	\caption{Test accuracy curves on \emph{MNIST} with different noise rates for all the compared methods. Colored curves show the mean accuracy of five trials, and shaded bars denote the standard deviations of the accuracies over five trials. }\vspace{-0.3cm}
	\label{fig1}
\end{figure*}

\begin{figure*}[!h]
	\begin{minipage}{1\linewidth}
		\centering
		{\includegraphics[width=0.28\linewidth]{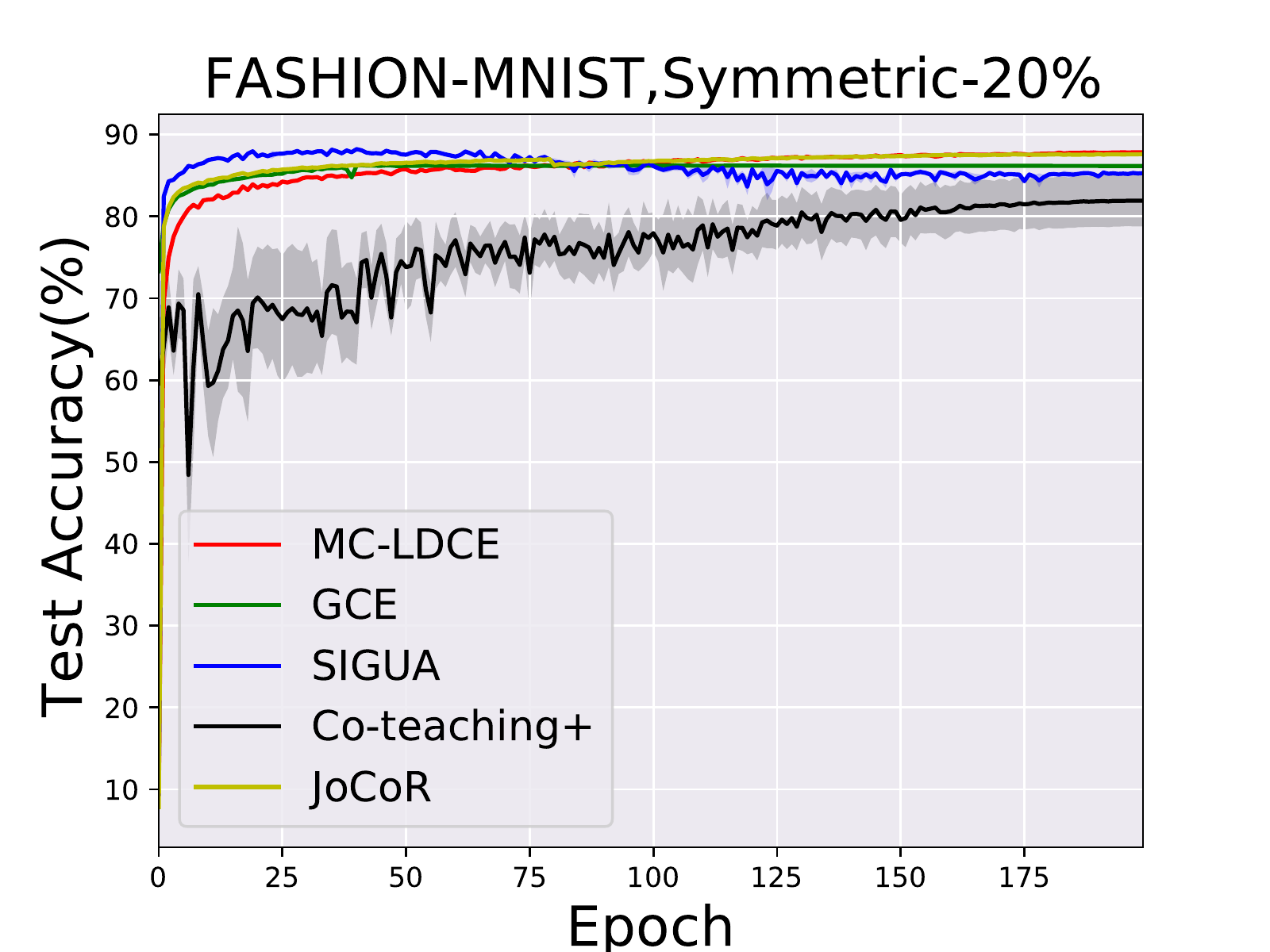}}		
		{\includegraphics[width=0.28\linewidth]{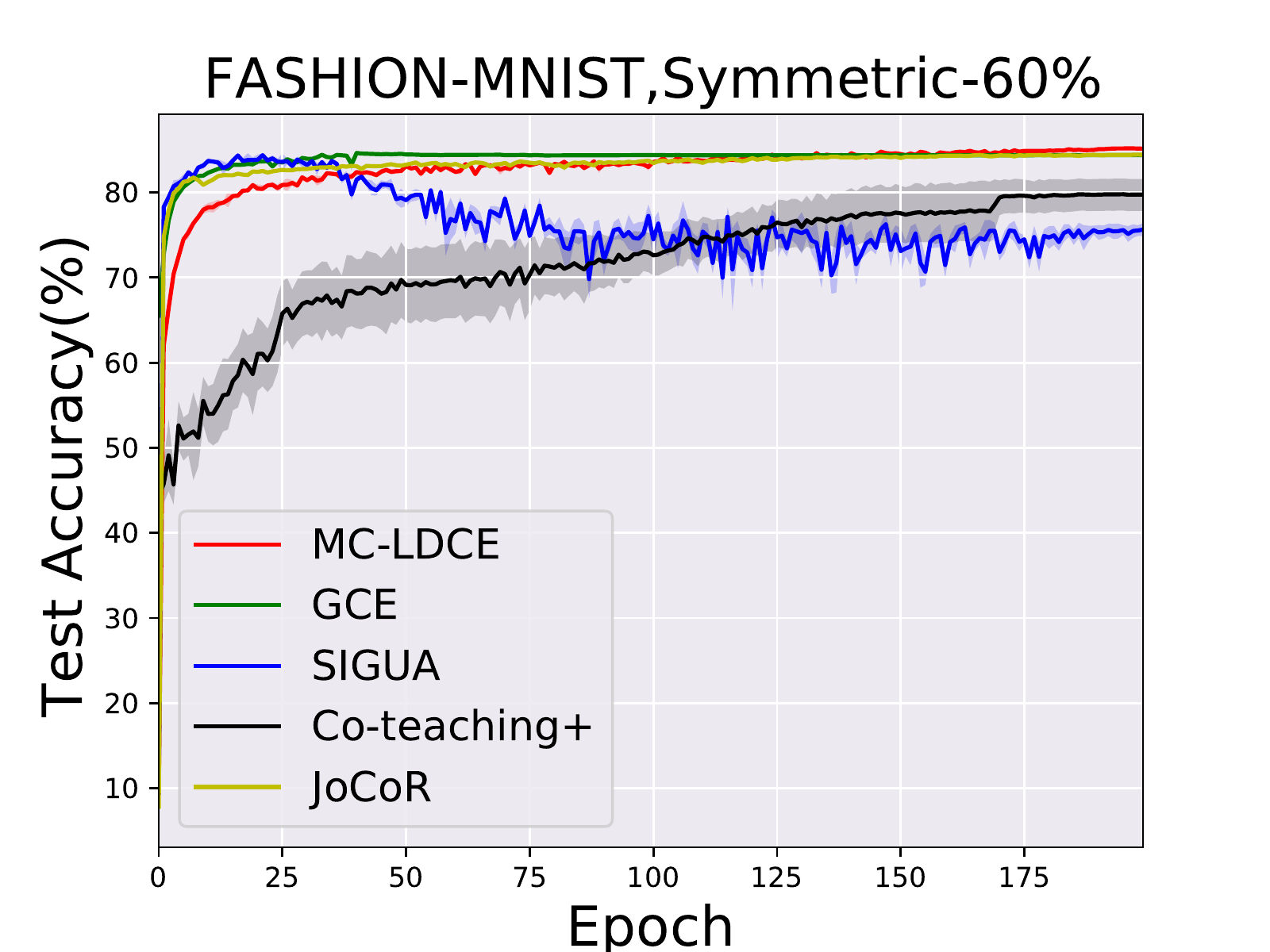}}
		{\includegraphics[width=0.28\linewidth]{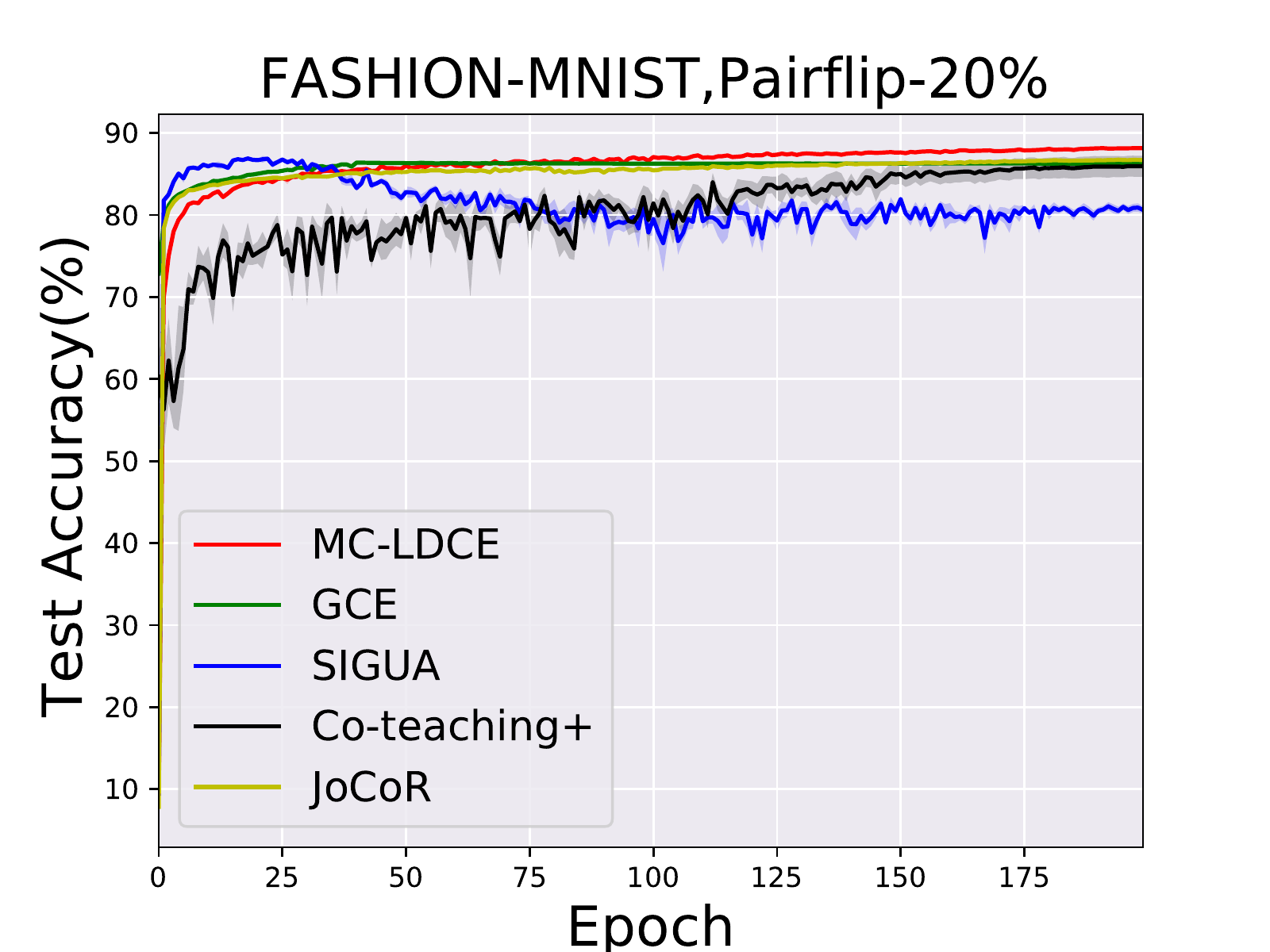}}
	\end{minipage} \vspace{-0.2cm}
	\caption{Test accuracy curves on \emph{FASHION-MNIST} with different noise rates for all the compared methods. Colored curves show the mean accuracy of five trials, and shaded bars denote the standard deviations of the accuracies over five trials. }\vspace{-0.35cm}
	\label{fig3}
\end{figure*}

\begin{figure*}[!h]
	\begin{minipage}{1\linewidth}
		\centering
		{\includegraphics[width=0.28\linewidth]{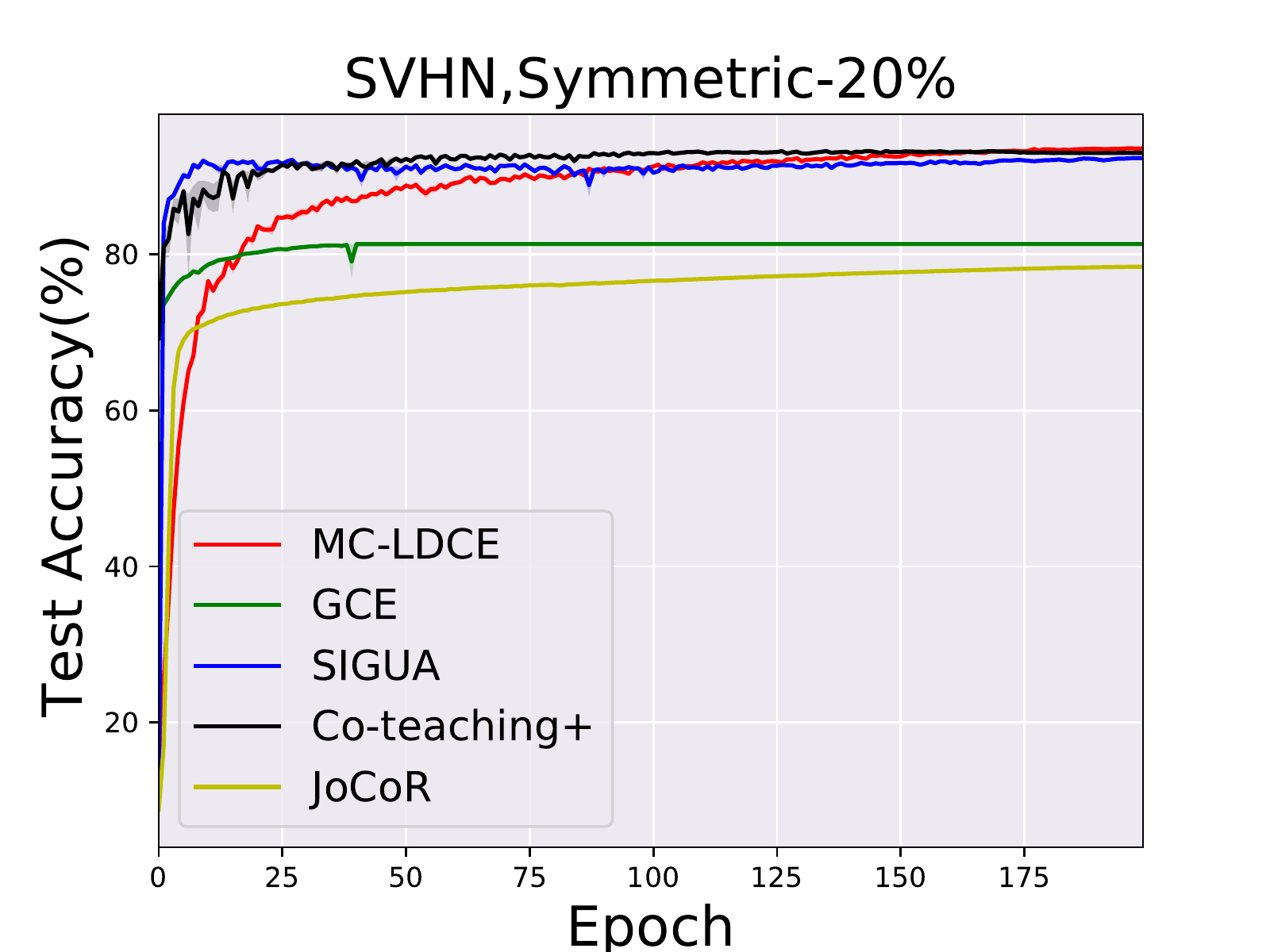}}		
		{\includegraphics[width=0.28\linewidth]{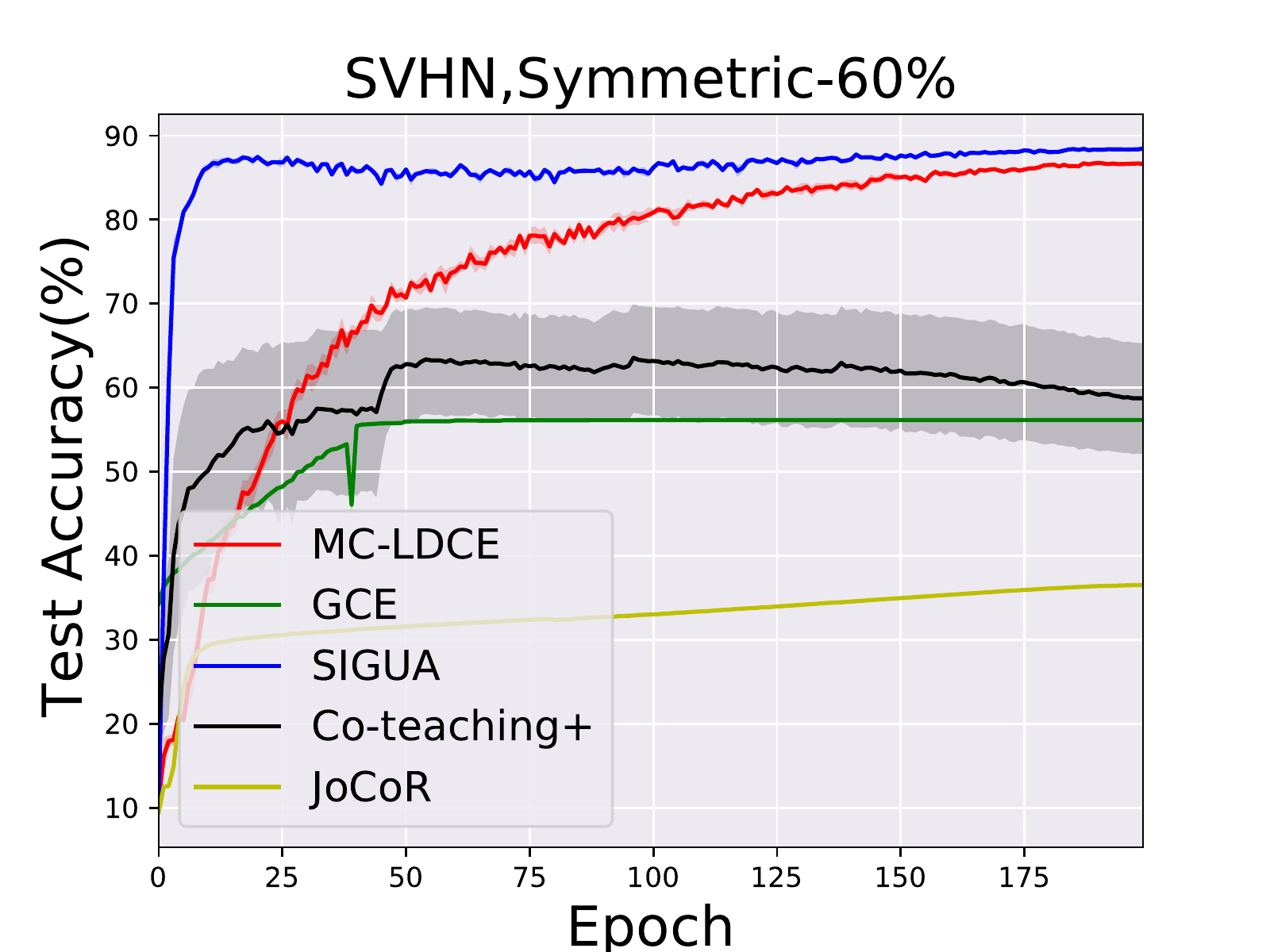}}
		{\includegraphics[width=0.28\linewidth]{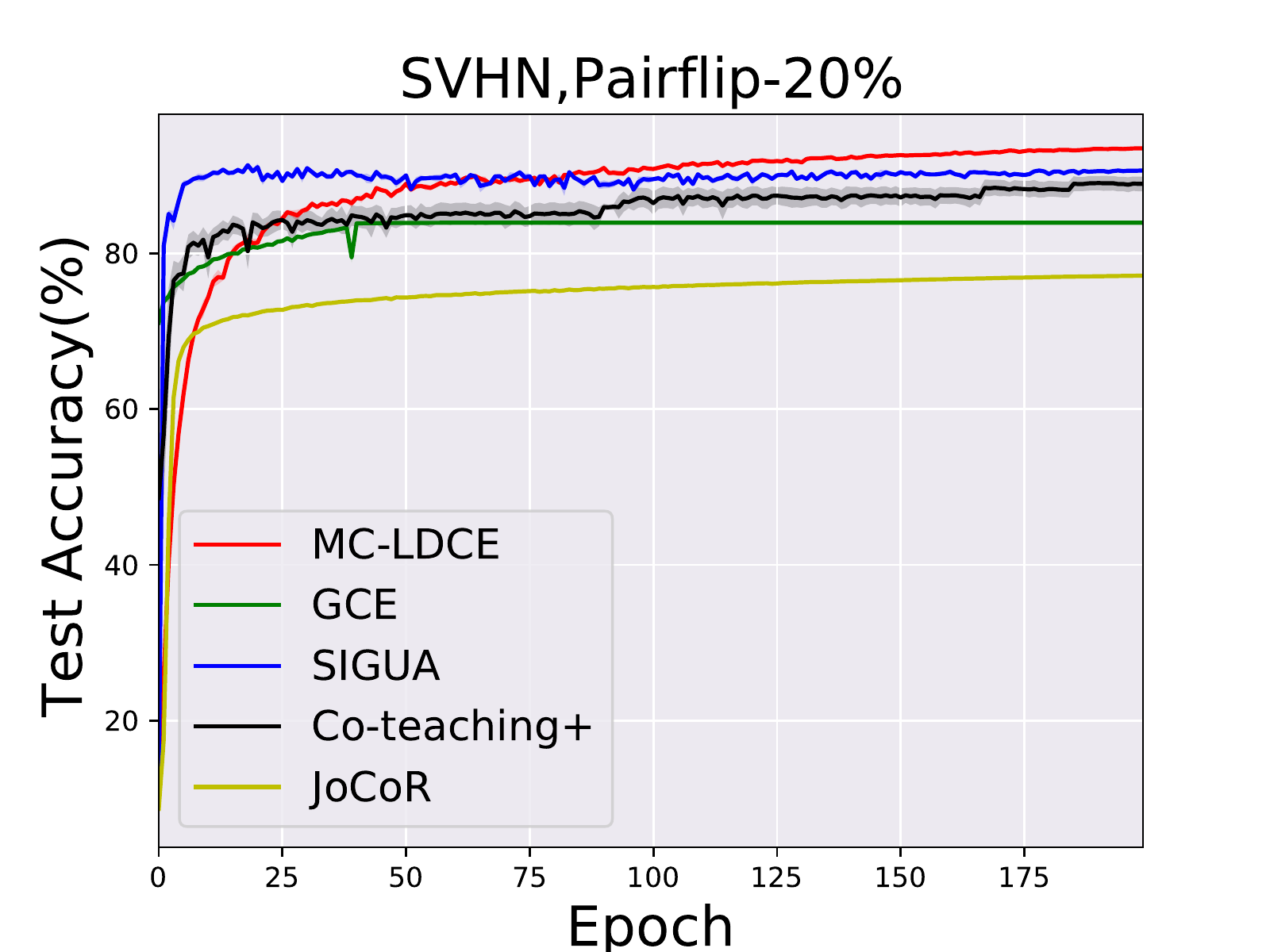}}
	\end{minipage} \vspace{-0.2cm}
	\caption{Test accuracy curves on \emph{SVHN} with different noise rates for all the compared methods. Colored curves show the mean accuracy of five trials, and shaded bars denote the standard deviations of the accuracies over five trials.} \vspace{-0.1cm}
	\label{fig4}
\end{figure*}


\paragraph{Real-world Noisy Dataset.}
\textit{Animal-10N} is introduced by \cite{song2019selfie} recently, which contains five pairs of confusing animals. The images are crawled from several online search engines including \textit{Bing} and \textit{Google} using the predefined labels as the search keyword. All label noise on \textit{Animal-10N} is introduced by human mistakes, and the overall noise rate on the training dataset is around $8\%$ while the test dataset is clean. This dataset contains $50,000$ RGB images used for training and $5,000$ RGB images for testing, and the resolution of each image is $64\times64$. 

\par

\paragraph{Compared Baselines.}
 We compare our MC-LDCE with several popular robust learning methods, including:

\begin{itemize}
	\setlength{\itemsep}{0pt}
	\setlength{\parsep}{0pt}
	\setlength{\parskip}{0pt}
	
	\item[$\bullet$] Co-teaching+ \cite{yu2019does} trains two deep neural networks simultaneously and lets them teach each other given every mini-batch.
	\item[$\bullet$] JoCoR \cite{wei2020combating} trains two networks and utilizes co-regularization to reduce the diversity of the two networks and combat the noisy labels.

	\item[$\bullet$] SIGUA \cite{han2020sigua} adopts gradient descent on ``good'' data while using a learning-rate-reduced gradient ascent on ``bad'' data.

	\item[$\bullet$] Generalized Cross-Entropy (GCE) \cite{zhang2018generalized} changes the loss function to make the trained neural network more robust in noisy label situations.

\end{itemize}

It is worth noting that we do not compare with other unbiased loss correction methods, such as ~\cite{natarajan2013learning,gao2016risk,patrini2016loss}, since they are not applicable to multi-class cases.

\par
\paragraph{Network Architectures.} We adopt a six-layer Convolutional Neural Network (CNN) as the backbone for \textit{CIFAR-10}, \textit{SVHN} and \textit{Animal-10N}, and two-layer Fully Connected Neural Network (MLP) for \textit{FASHION-MNIST} and \textit{MNIST}, which are both widely used in the related literature \cite{han2018co,wei2020combating,han2020sigua,malach2017decoupling}. The detailed network architectures are summarized in Table~\ref{tab:architecture}.

\par

\paragraph{Implementation Details.}
For a fair comparison, all experiments are conducted on an RTX2080-Ti GPU. The backbone network architectures are the same for all the methods, and we implement the compared baselines using their default parameters suggested by the original papers.
For our MC-LDCE, the model is trained over $200$ epochs, and we adopt the Adam algorithm to optimize our model with a momentum of $0.9$. The initial learning rate is set to $0.001$, and will be linearly decreased after $80$ epochs. The batchsize is set to $128$.

\subsubsection{Experimental Results} 
\par
\paragraph{Results on \textit{CIFAR-10}.} 
Figure~\ref{fig2} plots the test accuracy \textit{vs.} number of epochs on \textit{CIFAR-10}.
In the easiest Symmetry-$20\%$ case, the test accuracy of all compared methods increases steadily over the increase of epochs, which demonstrates their robustness. However, when meeting with a harder case, \ie, $60\%$ symmetric noise, Co-teaching+ and GCE first reach a very high level and then decrease gradually, which is because of the memorization effect of neural networks. To be specific, when the training proceeds, the neural network will tend to overfit the noisy examples which will lead to a decline in test accuracy. While the accuracy of our method increases steadily and finally exceeds all the others, verifying the robustness of our  MC-LDCE for extremely corrupted datasets (more than $50\%$ data are corrupted).
As for pairflip noise, we can see that our MC-LDCE outperforms the competitors with a large margin. For example, MC-LDCE exceeds the second best method with $1.83\%$ and $6.85\%$ in Pairflip-$20\%$ case and Pairflip-$40\%$ case, respectively. Thus, the proposed MC-LDCE is effective in dealing with both symmetric and pairflip label noise.
\par
\paragraph{Results on \textit{MNIST}.} 
For \textit{MNIST}, we evaluate the proposed method with synthetic label noise, \ie, symmetric label noise with the noise rate in $\{20\%, 60\%\}$ and pairflip label noise with the noise rate in $\{20\%, 40\%\}$.
We run five individual trials for all the compared methods under each noise level. 
Figure~\ref{fig1} (a) and (b) plot the test accuracy curves on \textit{MNIST} with $20\%$ and $60\%$ noise rates under symmetric label noise. Figure~\ref{fig1} (c) shows the test accuracy curves with the noise rate of $20\%$ under the pairflip label noise. 
Table~\ref{result_on_simulated_data} provides us the test accuracies and the corresponding standard deviations of all compared methods.
From the results, we can see that the accuracy of our MC-LDCE increases steadily over the increase of epochs, and our method outperforms other compared baselines finally, 
which indicates the effectiveness of our MC-LDCE in dealing with noisy labels.
\par
\paragraph{Results on \textit{FASHION-MNIST}.} 
Figure~\ref{fig2} shows the test accuracy \textit{vs.} number of epochs on \textit{FASHION-MNIST}. 
Similarly, the accuracy of our MC-LDCE grows steadily over the increase of epochs and outperforms the other compared baseline gradually. As shown in Table~\ref{result_on_simulated_data}, our method on \textit{FASHION-MNIST} consistently outperforms all the compared methods on all label noise cases, which demonstrate the superiority of the proposed method.
\par
\paragraph{Results on \textit{SVHN}.} The comparison results on \textit{SVHN} with different types of noise and different noise rates are shown in Figure~\ref{fig4} and  Table~\ref{result_on_simulated_data}.
From the comparison results, it can be seen that our MC-LDCE 
grows stably with the epoch increasing and gradually outperforms the compared methods with a large margin, especially for JoCoR, GCE, and Co-teaching+. In addition, as shown in Table~\ref{result_on_simulated_data}, our MC-LDCE can consistently achieve the best or the second best performance among all the compared methods. It is noted that the additional experimental results on four simulated noisy datasets can be found in the \textbf{supplementary material}.
\par
\paragraph{Results on \textit{Animal-10N}.}
Similar to the experimental settings on \textit{CIFAR-10}\footnote{The only difference lies in the input dimension of the last fully connected layer of the network architecture, where 1024 is used for \textit{Animal-10N}, while 512 is for \textit{CIFAR-10}, as the sizes of their images are different.}, we run five individual trials for all compared methods on \textit{Animal-10N}. Note that we do not apply any data augmentation or pre-processing procedures. Table ~\ref{result_on_Animal_10N} shows the average test accuracies and corresponding standard deviations of all compared methods on \textit{Animal-10N}, where we can see that our MC-LDCE achieves the highest classification accuracy among all comparators. Therefore, the proposed MC-LDCE is effective in handling real-world label noise.

\subsection{Experimental Results with Linear Model} 
\hspace*{\fill} \\
\noindent

  \renewcommand\arraystretch{0.4}
	\begin{table}[t]
		\centering
		\setlength\tabcolsep{23pt}
		\caption{Average test accuracy on \emph{Animal-10N}. The best results are marked in \textbf{bold}.} \vspace{-0.1cm}
			\begin{tabular}{c|c}
				\toprule[1.0pt]
			Method 		& Accuracy ($\%$) \\ 
			\toprule[1.0pt]
				 GCE~\cite{zhang2018generalized}   &  68.7 $\pm$ 0.04  \\
				 Co-teaching+~\cite{yu2019does}    & 69.7 $\pm$ 0.11 \\
				 JoCoR~\cite{wei2020combating}     & 75.7 $\pm$ 0.12  \\
				 SIGUA~\cite{han2020sigua}         & 74.0 $\pm$ 0.21 \\
				 MC-LDCE                           & \textbf{76.6 $\pm$ 0.23}  \\ 
				\toprule[0.8pt]
			\end{tabular}
			\label{result_on_Animal_10N}
    \end{table}

In this part, we equip our MC-LDCE with a linear classification model and compare it with several statistical learning-based robust methods.
The compared methods include: 1) Unbiased Logistic Estimator (ULE)~\cite{natarajan2013learning}, 2) $\mu$ Stochastic Gradient Descent ($\mu$SGD) \cite{patrini2016loss}, 3) Spectral Cluster Discovery (SCD) \cite{luo2020bi}, 4) Rank Pruning (RP) \cite{northcutt2017learning}, and 5) Label Noise handling via Side Information (LNSI) \cite{wei2019harnessing}. Note that the first two approaches are originally designed for binary classification tasks, so we use the one-vs-all strategy to apply them to multi-class cases. Details of all the compared methods can be found in the \textbf{supplementary materials}.
In the experiments, we evaluate the proposed method on corrupted \textit{CIFAR-10}. To be specific, 
we randomly pick up $30,000$ images from \textit{CIFAR-10} across different classes and corrupt them with different levels of symmetric noise. 
The classification accuracies of all the compared approaches under different noise levels are shown in Table~\ref{result_on_linear_model}. It is worth noting that the proposed MC-LDCE consistently outperforms all the competitors under various noise levels, which again demonstrates the superiority of MC-LDCE in dealing with label noise.

\renewcommand\arraystretch{0.4}
\begin{table}[t]
	\centering
	\caption{Average test accuracy on \emph{CIFAR-10} using a linear classification model. The best results are marked in \textbf{bold}.} \vspace{-0.1cm}
	\begin{tabular}{r|c|c|c}
		\toprule[1.0pt]
	Method 					& 20\% 	                 & 40\% 		  & 60\%  		          \\ 
	\toprule[1.0pt]
		 ULE~\cite{natarajan2013learning}  &  74.8$\pm$0.050 & 61.7$\pm$0.075 & 41.5$\pm$0.068 \\
		 $\mu$SGD~\cite{patrini2016loss}    & 74.1$\pm$0.009 & 72.6$\pm$0.012 & 71.6$\pm$0.001 \\
		 RP~\cite{northcutt2017learning}    & 77.9$\pm$0.015 & 64.6$\pm$0.009 & 47.4$\pm$0.006 \\
		 LNSI~\cite{wei2019harnessing}  & 84.7$\pm$0.006 & 83.8$\pm$0.006 & 77.4$\pm$0.006 \\
		 SCD~\cite{luo2020bi}   & 86.5$\pm$0.007 & 84.5$\pm$0.006 & 77.6$\pm$0.020 \\
		 MC-LDCE   
		   & \textbf{87.1$\pm$0.359} & \textbf{85.1$\pm$0.441} & \textbf{79.7$\pm$0.884} \\ 
	\toprule[0.8pt]
	\end{tabular}
	\label{result_on_linear_model}
\end{table}

\section{Conclusion}

In this paper, we propose a novel multi-class LNL method termed MC-LDCE to deal with the label noise problem. In the proposed MC-LDCE, we decompose the multi-class classification loss (\eg, mean squared loss) into label-independent and label-dependent parts, and directly estimate the label-dependent part via centroid estimation. Our MC-LDCE is the first method based on LDCE to deal with multi-class LNL problems. Furthermore, as our MC-LDCE is independent of the classification model, we conduct intensive experiments by using deep and linear models on both simulated and real-world noisy datasets. Experimental results demonstrate that our MC-LDCE outperforms other representative LNL methods. 

\noindent\textbf{Acknowledgement}. This research is supported by NSF of China (Nos: 61973162, 62172228), NSF of Jiangsu Province (No: BZ2021013), and the Fundamental Research Funds for the Central Universities (Nos: 30920032202, 30921013114), the GuangDong Basic and Applied Basic Research Foundation (No: 2020A1515110554), and the Science and Technology Program of Guangzhou, China (No: 202002030138).

\bibliographystyle{spbasic}      


\bibliography{my}

\end{document}